%% file: main.tex
\documentclass[singlecolumn,11pt]{article}
\usepackage{fullpage}
\usepackage[svgnames,dvipsnames,table]{xcolor}
\usepackage{hyperref}
\hypersetup{colorlinks,linkcolor={blue},citecolor={blue},urlcolor={RoyalBlue}}

\usepackage{amsthm}
\usepackage{amsmath} 
\usepackage{nicefrac}
\usepackage{bm}
\usepackage{booktabs}
\usepackage{amsfonts}
\usepackage{bbm}
\usepackage{mleftright}
\usepackage{dsfont}
\usepackage{amscd,amssymb,amsmath,amsthm,bbold}
\usepackage{graphicx}
\usepackage{stfloats}
\usepackage{multirow}
\usepackage{wrapfig}
\usepackage[numbers]{natbib}

\usepackage{pifont}
\usepackage{microtype}
\usepackage{enumitem}
\usepackage{xfrac}
\usepackage{parskip}
\usepackage{numprint}
\usepackage{booktabs} 

\usepackage{transparent}

\usepackage{algorithm}
\usepackage{algorithmic}
\usepackage{listings}
\usepackage{xspace}
\usepackage{subcaption}
\usepackage{etoolbox}
\usepackage{comment}
\usepackage{etoc}
\usepackage{wrapfig}
\usepackage{placeins}
\usepackage{makecell}
\usepackage{lipsum}

\usepackage{microtype}
\usepackage{booktabs} 
\usepackage{makecell}

\usepackage{amsmath}
\usepackage{amssymb}
\usepackage{mathtools}
\usepackage{amsthm}
\usepackage{chngcntr}

\usepackage{nicefrac}
\usepackage{bm}
\usepackage{booktabs,tabularx}
\usepackage{amsfonts}
\usepackage{bbm}
\usepackage{mleftright}
\usepackage{dsfont}
\usepackage{placeins}
\usepackage{amscd,amssymb,amsmath,amsthm,bbold}
\usepackage[normalem]{ulem}
\usepackage[font=small]{caption}
\usepackage{bbm}
\usepackage{rotating}
\usepackage{tcolorbox}
\usepackage{pgfmath}
\usepackage{float}

\usepackage{epigraph}
\usepackage{pifont}

\newcommand{\cmark}{\textcolor{green}{\ding{51}}}
\newcommand{\xmark}{\textcolor{red}{\ding{55}}}

\usepackage{adjustbox}
\usepackage{array}
\usepackage{soul}

\newcolumntype{R}[2]{%
    >{\adjustbox{angle=#1,lap=\width-(#2)}\bgroup}%
    l%
    <{\egroup}%
}

\theoremstyle{plain}

\theoremstyle{definition}

\theoremstyle{remark}

\mathchardef\mhyphen="2D

\usepackage[textsize=tiny]{todonotes}
\NewDocumentCommand\kaleemoji{}{\includegraphics[width=0.35cm]{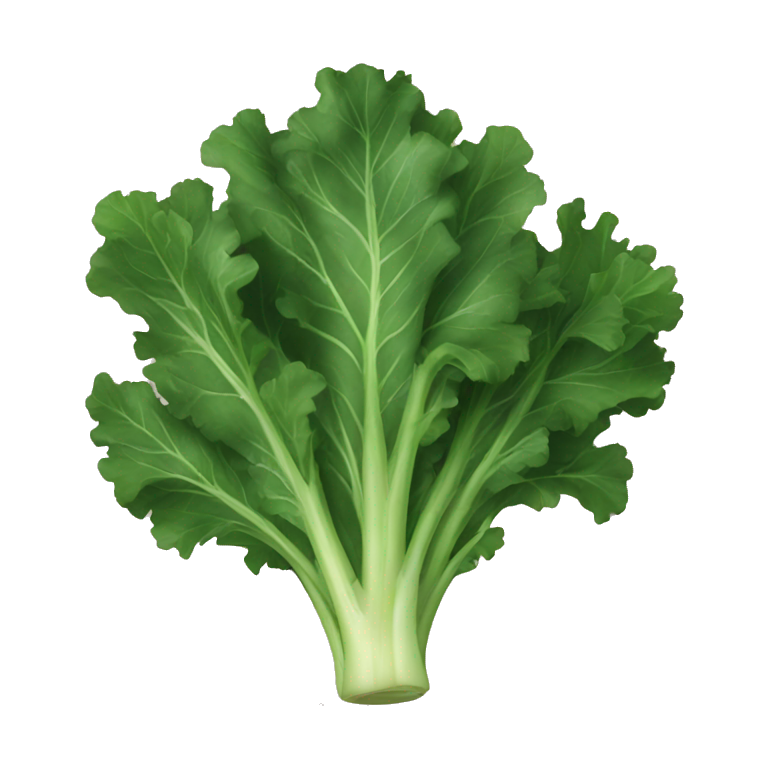}}

\begin{document}

\date{}
\title{\includegraphics[width=1cm]{figs/kale.png}\Large BLIP3-KALE: Knowledge Augmented Large-Scale Dense Captions}

\author{%
  \small\textbf{Anas Awadalla$^{1,2}$}\thanks{Work done while interning at Salesforce Research} \quad
  \textbf{Le Xue$^{2}$} \quad
  \textbf{Manli Shu$^{2}$} \quad
  \textbf{An Yan$^{2}$} \quad
  \textbf{Jun Wang$^{2}$} \quad
  \textbf{Senthil Purushwalkam$^{2}$} \quad
  \\
  \small\textbf{Sheng Shen$^{4}$} \quad
  \textbf{Hannah Lee$^{1}$} \quad
  \textbf{Oscar Lo$^{1}$} \quad
  \textbf{Jae Sung Park$^{1}$} \quad
  \textbf{Etash Guha$^{1}$} \quad
  \\
  \small\textbf{Silvio Savarese$^{\diamond,2,3}$} \quad
  \textbf{Ludwig Schmidt$^{\diamond,1}$} \quad
  \textbf{Yejin Choi$^{\diamond,1}$} \quad
  \textbf{Caiming Xiong$^{\diamond,2}$} \quad
  \textbf{Ran Xu$^{\diamond,2}$} \quad
  \\
  \footnotesize{$^1$ University of Washington, $^2$ Salesforce Research, $^3$ Stanford University,} \\
  \footnotesize{$^4$ University of California, Berkeley, ${^\diamond}${Senior Authors}}
}

\maketitle
\vspace{-0.5cm}
\begin{abstract}
We introduce \kaleemoji BLIP3-KALE, a dataset of 218 million image-text pairs that bridges the gap between descriptive synthetic captions and factual web-scale alt-text. KALE augments synthetic dense image captions with web-scale alt-text to generate factually grounded image captions. Our two-stage approach leverages large vision-language models and language models to create knowledge-augmented captions, which are then used to train a specialized VLM for scaling up the dataset. We train vision-language models on KALE and demonstrate improvements on vision-language tasks. Our experiments show the utility of KALE for training more capable and knowledgeable multimodal models. We release the KALE dataset at \url{https://huggingface.co/datasets/Salesforce/blip3-kale}.
\end{abstract}

\begin{figure*}[!ht]
    \centering
    \includegraphics[width=\linewidth]{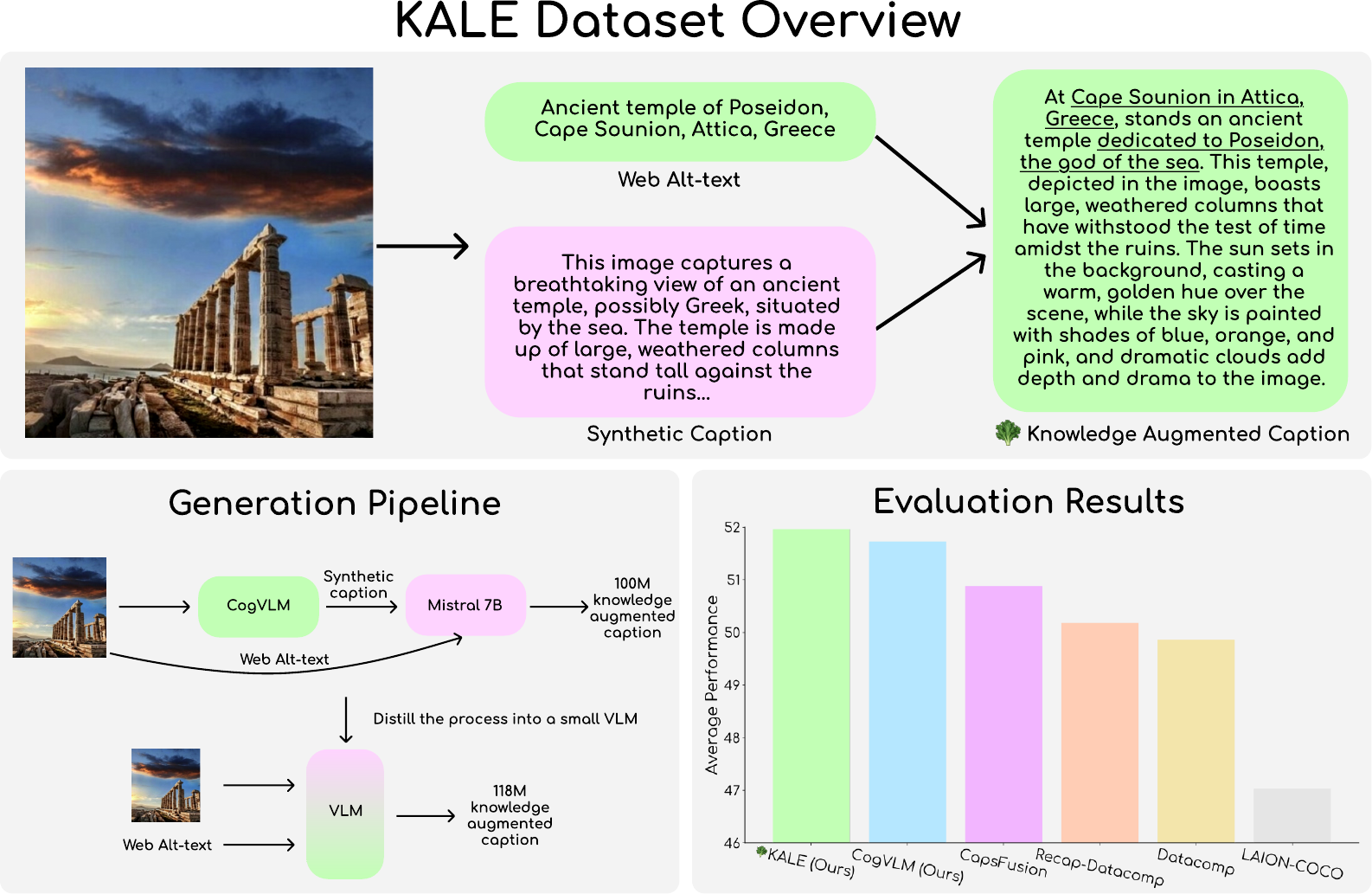}
    \caption{Overview of \kaleemoji KALE dataset creation and performance. \textbf{Top:} Example showing how KALE combines web alt-text with synthetic captions to produce knowledge-rich dense captions. \textbf{Bottom left:} Two-stage generation pipeline for KALE, using CogVLM and Mistral to create an initial set of knowledge augmented captions, followed by training a distilled VLM to scale up to 218M samples. \textbf{Bottom right:} Evaluation results comparing KALE's average performance against popular synthetic image-text datasets.}
    \label{fig:overview}
\end{figure*}
\footnotetext[1]{\url{https://laion.ai/blog/laion-coco/}}

\input{sections/introduction}

\input{sections/approach}

\input{sections/experiments}

\input{sections/related_works}

\input{sections/conclusion}

{\small
\bibliographystyle{plainnat}
\bibliography{main}
}
\clearpage
\appendix

\end{document}

%% file: sections/introduction.tex
\section{Introduction}
\begin{table}[h]
\centering
\resizebox{\textwidth}{!}{%
\begin{tabular}{@{}lcccc@{}}
\toprule
Dataset & Scale (\# of samples) & Density (avg. words/caption) & Knowledge-augmented? & Captioner size (params)\\
\midrule
LAION-COCO$^1$ & 600M & 8.99 & \xmark & 0.5B\\
ReCap-Datacomp-1B~\cite{Li2024WhatIW} & 1.28B & 49.43 & \xmark & 7B\\
CapsFusion~\cite{Yu2023CapsFusionRI} & 120M & 22.74 & \cmark & 0.5B\\
\kaleemoji KALE & 218M & 67.26 & \cmark & 17B (stage 1) $\rightarrow$ 2B (stage 2)\\
\bottomrule
\end{tabular}}
\caption{\textbf{Comparison of open-source synthetic image-text datasets:} We compare various datasets in terms of scale (number of samples), density (average number of words per sample), whether they are knowledge-augmented (meaning that the caption includes information found in image's web scraped alt-text), and the size of the captioning model used to generate the descriptions. For KALE, we create an initial pool of 100M captions from a 17B parameter model and use it to distill a 2B parameter model that matches the performance of the larger 17B model.}
\label{tab:image-text-datasets}
\end{table}

We introduce BLIP3-KALE, a dataset of 218 million image-text pairs that advances the state of knowledge-augmented image captioning. KALE builds upon recent work in this area, particularly CapsFusion~\cite{Yu2023CapsFusionRI}, which pioneered the use of large language models to fuse synthetically generated captions with alt-text to incorporate real-world knowledge.
KALE makes two key contributions beyond CapsFusion:

\textbf{Scale and Density:} While CapsFusion produced 120M samples with an average of 22.74 words per caption, KALE is significantly larger and denser. It contains 218M samples with an average of 67.26 words per caption - 1.82x the scale and nearly 3x the density of CapsFusion.

\textbf{Efficient Generation:} We distill the knowledge augmentation process into a compact 2B parameter captioning model. This enables generation of high-quality captions comparable to much larger models like CogVLM-17B~\cite{Wang2023CogVLMVE}, but at a fraction of the cost. This efficiency allows us to scale up the dataset creation process.

Our approach combines synthetic captions from VLMs with factual information from web-scale alt-text, creating rich image descriptions. We demonstrate that training on KALE improves performance across multimodal tasks compared to many previous purely synthetic or web-scraped datasets.

%% file: sections/approach.tex
\section{Approach}

\begin{figure*}[thb]
    \centering
    \includegraphics[width=\linewidth]{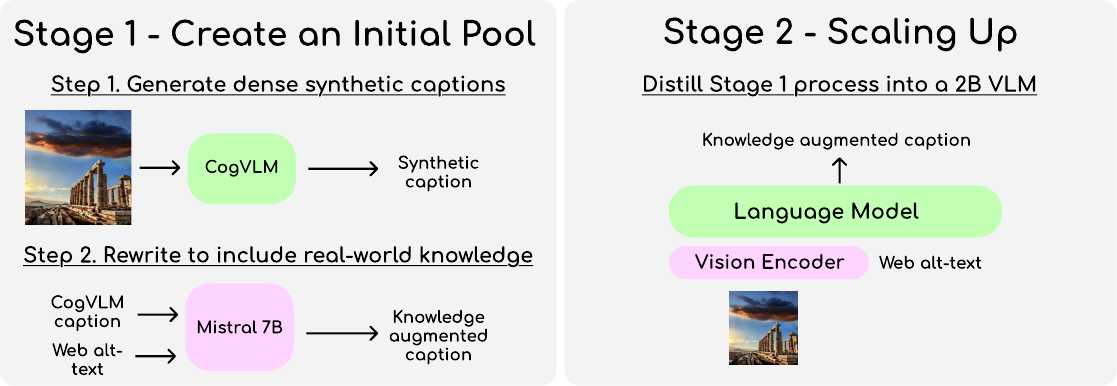}
    \caption{We generate \kaleemoji KALE in a two stage process. \textbf{Stage 1:} We first create an initial pool of 100M knowledge-augmented dense captions using CogVLM-17B to generate dense captions for Datacomp-1B images and then augmenting these captions with real world knowledge by prompting Mistral. \textbf{Stage 2:} We use the knowledge-augmented captions from Stage 1 to train a VLM that takes image patch embeddings and Datacomp-1B captions as inputs and outputs knowledge-augmented captions. This VLM is then used to efficiently caption 118M more images from Datacomp-1B.
    }
    \label{fig:approach}
\end{figure*}

\subsection{Stage 1: Generating initial knowledge-augmented captions}
The first stage of our approach focuses on creating an initial pool of knowledge-augmented dense captions. We begin by leveraging CogVLM-17B~\cite{Wang2023CogVLMVE} to generate dense captions for images from the Datacomp-1B dataset. These captions serve as a foundation for our knowledge augmentation process.
To enhance these captions with real-world knowledge, we employ Mistral, a powerful language model.We prompt Mistral using the CogVLM-generated captions, instructing it to augment the descriptions with relevant factual information. This step aims to incorporate broader contextual knowledge into the image descriptions, following CapsFusions' prompting method. Through this process, we create an initial pool of 100 million knowledge-augmented captions.

\subsection{Stage 2: Scaling up}
The second stage of our approach focuses on scaling up the dataset to achieve our target of 218 million image-text pairs. We accomplish this by training a specialized VLM using the knowledge-augmented captions generated in Stage 1. We construct our VLM similar to the LLaVA~\cite{Liu2023VisualIT} model; using Qwen1.5-1.5B~\cite{Yang2024Qwen2TR} for the language model and DFN ViT-H~\cite{fang2023data} for the vision encoder. We resize the positional embeddings for the vision encoder to handle 490x490 images matching the resolution used by CogVLM.
Our VLM takes two inputs: image patch embeddings and the original Datacomp-1B captions. The model is trained to output the knowledge-augmented captions produced in Stage 1.
We use this VLM to caption an additional 118 million images from the Datacomp-1B dataset. This two-stage approach enables us to efficiently scale up KALE to 218 million image-text pairs.

\begin{wrapfigure}{r}{0.5\textwidth}
\vspace{-0.1cm}
\centering
\fbox{
\begin{minipage}{0.95\linewidth}
\centering

\vspace{0.3cm}
\footnotesize
The Quality and project coordinator position at Credit Agricole (Dublin), 
\textcolor{red}{\textbf{as indicated in Sentence 1}}, is a part of a VIE (Venture in Ireland), 
and \textcolor{red}{\textbf{my analysis now turns to the visual representation of this organization}}. 
The logo, \textcolor{red}{\textbf{as depicted in Sentence 2}}, features a stylized green 'A' letter, 
which is accentuated by blue and red lines on the right side. The dynamic and modern 
appearance is achieved by intertwining the 'A' with the blue and red line. 
A horizontal green line lies beneath the letter 'A'.
\vspace{0.3cm}
\end{minipage}
}
\caption{Example of pipeline artifacts in a caption. The highlighted text in red shows phrases that have leaked from the system prompt into the final output.}
\label{fig:pipeline_artifact}
\vspace{-1cm}
\end{wrapfigure}

\subsection{Removing pipeline artifacts}
Artifacts from the prompts passed in to LLMs/VLMs to generate KALE occasionally leak into the generated captions. We present an example of these artifacts in Figure~\ref{fig:pipeline_artifact} where the system prompt to the LLM used in the rewriting stage has leaked into the generated caption. To remove these artifacts we create a set of words that commonly appear in these artifacts such as `real-world` or `sentence structure` and remove sentences that contain these keywords.

%% file: sections/experiments.tex
\section{Experiments}
We validate the effectiveness of KALE by training VLMs. In this section we outline our training and evaluation setup and present results when training on KALE.

\subsection{Training setup}
We follow the Llava architecture in using a linear layer to project the image patch embeddings from a vision encoder into the text embeddings space. We use Qwen2.5-1.5B~\cite{Yang2024Qwen2TR} for the language model, and SigLIP ViT-L 384~\cite{Zhai2023SigmoidLF} for the vision encoder. We use a batch size of 80 image-text pairs and a peak learning rate of $5e^{-5}$. We train all of our models on two million samples from image-text data. We then fine-tune the model, using a peak learning rate of $3e^{-5}$, on one million multimodal instruction tuning samples from the Cauldron~\cite{Laurenon2024WhatMW} dataset. We remove multi-image samples from the Cauldron data and sample different subsets according to the ratios used in Idefics2. 

\subsection{Evaluation setup}
We evaluate the instruction-tuned model on various vision-language benchmarks including TextVQA (val set)~\cite{singh2019towards}, VQAv2 (val lite)~\cite{Agrawal2015VQAVQ}, ScienceQA~\cite{lu2022learn}, AI2D~\cite{Kembhavi2016ADI}, MMBench~\cite{Liu2023MMBenchIY}, ChartQA~\cite{Masry2022ChartQAAB}, InfoVQA~\cite{Mathew2021InfographicVQA}, OCRBench~\cite{liu2024ocrbenchhiddenmysteryocr}, RealWorldQA\footnote{\url{https://x.ai/blog/grok-1.5v}}, and MMStar~\cite{chen2024we} using the lmms-eval framework~\cite{zhang2024lmmsevalrealitycheckevaluation}. This comprehensive evaluation suite covers a wide range of capabilities from general visual question answering to specialized tasks involving scientific reasoning and OCR-based comprehension.

\subsection{Results}
\begin{table}[h]
\centering
\resizebox{\textwidth}{!}{%
\begin{tabular}{lccccccccccr}
\toprule
Model & \multicolumn{11}{c}{Benchmarks} \\
\cmidrule(lr){2-12}
& TextVQA & VQAv2 & ScienceQA & AI2D & MMBench & ChartQA & InfoVQA & OCRBench & RealWorldQA & MMStar & Avg \\
\midrule
\kaleemoji KALE {\footnotesize\textbf{(Ours)}} & \textbf{59.92} & \textbf{70.10} & \textbf{72.68} & \textbf{65.64} & 58.59 & 23.28 & 29.28 & \textbf{43.80} & 52.42 & \textbf{43.91} & \textbf{51.96} \\
CogVLM {\footnotesize\textbf{(Ours)}} & 59.74 & 69.42 & 70.30 & 65.35 & \textbf{61.60} & \textbf{23.64} & \textbf{29.53} & \textbf{43.80} & 52.03 & 41.90 & 51.73 \\
CapsFusion & 57.62 & 67.30 & 71.79 & 62.27 & 60.82 & 22.28 & 27.67 & 43.10 & 52.03 & \textbf{43.91} & 50.88 \\
Recap-Datacomp & 58.49 & 67.36 & 71.19 & 63.31 & 52.75 & 23.08 & 28.45 & 42.20 & \textbf{53.07} & 41.90 & 50.18 \\
Datacomp & 57.40 & 67.22 & 69.51 & 61.82 & 59.45 & 22.28 & 28.53 & 42.20 & 50.46 & 39.70 & 49.86 \\
LAION-COCO & 54.12 & 65.26 & 65.94 & 59.10 & 55.58 & 21.60 & 26.81 & 38.90 & 44.05 & 38.90 & 47.03 \\
\bottomrule
\end{tabular}}
\caption{\textbf{Downstream performance:} To measure the quality of KALE in comparison to other datasets, we evaluate the instruction-tuned models across vision-language tasks. KALE maintains a slight edge in overall performance, while our CogVLM synthetic captions shows strong performance in tasks like MMBench. Both subsets of our KALE data outperform existing synthetic image-text datasets.}
\label{tab:pre-training-performance}
\end{table}

We find that pre-training on KALE captions improve downstream model performance on most VLM benchmarks, achieving the highest average performance at 51.96\%. In particular, KALE shows strong performance on TextVQA (59.92\%), VQAv2 (70.10\%), and ScienceQA (72.68\%). CogVLM's synthetic captions also demonstrate robust performance. Both KALE and CogVLM significantly outperform Datacomp-1B's noisier alt-text captions, which achieves lower scores across most benchmarks (49.86\% average). Earlier attempts at knowledge integration, such as CapsFusion (50.88\% average), while showing improvements over the Datacomp baseline, didn't achieve the same level of performance as our approach. The LAION-COCO dataset, constrained by both vocabulary size and caption density, performs the lowest at 47.03\% average. Furthermore, Table~\ref{tab:different-stages-model-comparisons} compares the performance of stage 1 captions generated by CogVLM and Mistral-7B with the complete KALE dataset, which combines these stage 1 captions with those from our distilled captioning model (stage 2). The combined stage 1 and 2 captions achieve comparable performance to the stage 1 captions generated by the significantly larger CogVLM model, demonstrating the effectiveness of our distilled captioning approach.

\begin{wraptable}{r}{0.3\textwidth}
\centering
\small{\begin{tabular}{lc}
\toprule
Model & Average \\
\midrule
KALE \scriptsize{(stage 1 only)} & 51.53 \\
\kaleemoji KALE & 51.96 \\
\bottomrule
\end{tabular}}
\caption{Average performance shows little difference between training on stage 1 captions and a mixture of stage 1 and stage 2 (i.e.\kaleemoji KALE).}
\label{tab:different-stages-model-comparisons}
\end{wraptable}

%% file: sections/related_works.tex
\section{Related Works}
KALE builds on many large-scale image-text datasets such as LAION-5B~\cite{Schuhmann2022LAION5BAO}, Datacomp-1B~\cite{datacomp}, COYO-700M~\cite{kakaobrain2022coyo-700m}, and many more. These datasets were sourced from large amounts of images paired with alt-text captions found in the HTML image tags associated with these images. As LLM/VLMs have become more capable, many works have explored generating synthetic multimodal training data. There is a line of work that seeks to improve image-text datasets by using VLMs to generate synthetic captions~\citep{Nguyen2023ImprovingMD,Li2022BLIPBL,Li2024WhatIW,li2024llavanext-ablations}. Works such as LaCLIP~\cite{fan2023improving} took an alternative approach of rewriting the existing alt-text caption using an LLM to improve caption quality. Moreover, works such as LLaVA~\cite{Liu2023VisualIT} and LLaVAR~\cite{Zhang2023LLaVAREV} have synthetically generated visual question-answer pairs in the context of instruction tuning data. Additionally the xGen-mm~\cite{xgen_mm_phi3_mini} model leverages our KALE dataset to improve the quality of their caption data.

Previous work has pointed out that synthetic captions lack real-world knowledge, limiting their applicability in many domains. CapsFusion~\cite{Yu2023CapsFusionRI} addresses this issue by augmenting LAION-COCO synthetic captions with alt-text from the LAION dataset. VeCLIP~\cite{Lai2023VeCLIPIC} also addresses this issue but instead of using existing captions, it generates synthetic captions using a LLaVA model.

An adjacent line of work improves the text quality of multimodal data by instead sourcing web-scale interleaved image/text samples from web documents as opposed to HTML alt-text captions. Works such as MMC4~\cite{Zhu2023MultimodalCA}, OBELISC~\cite{Laurenccon2023OBELISCAO}, MINT-1T~\cite{Awadalla2024MINT1TSO}, and OmniCorpus~\cite{Li2024OmniCorpusAU} all build multimodal interleaved datasets, which is a promising direction for attaining high-quality and knowledge rich multimodal data.

%% file: sections/conclusion.tex
\section{Limitations and conclusion}
KALE represents a step forward in bridging the gap between descriptive synthetic captions and factual web-scale alt-text. Our experiments demonstrate that models trained on KALE consistently outperform baseline models across various benchmarks. While KALE performs favorably compared to other open-source image-text datasets, the data still suffers from hallucination, particularly in text-dense images. Future work should scale KALE to billions of image-text pairs, explore more sophisticated knowledge augmentation techniques, and investigate its impact on a broader range of multimodal tasks.

%% file: main.bbl
\begin{thebibliography}{32}
\providecommand{\natexlab}[1]{#1}
\providecommand{\url}[1]{\texttt{#1}}
\expandafter\ifx\csname urlstyle\endcsname\relax
  \providecommand{\doi}[1]{doi: #1}\else
  \providecommand{\doi}{doi: \begingroup \urlstyle{rm}\Url}\fi

\bibitem[Agrawal et~al.(2015)Agrawal, Lu, Antol, Mitchell, Zitnick, Parikh, and Batra]{Agrawal2015VQAVQ}
Aishwarya Agrawal, Jiasen Lu, Stanislaw Antol, Margaret Mitchell, C.~Lawrence Zitnick, Devi Parikh, and Dhruv Batra.
\newblock Vqa: Visual question answering.
\newblock \emph{International Journal of Computer Vision}, 123:\penalty0 4 -- 31, 2015.
\newblock URL \url{https://api.semanticscholar.org/CorpusID:3180429}.

\bibitem[Awadalla et~al.(2024)Awadalla, Xue, Lo, Shu, Lee, Guha, Jordan, Shen, Awadalla, Savarese, Xiong, Xu, Choi, and Schmidt]{Awadalla2024MINT1TSO}
Anas Awadalla, Le~Xue, Oscar Lo, Manli Shu, Hannah Lee, Etash~Kumar Guha, Matt Jordan, Sheng Shen, Mohamed Awadalla, Silvio Savarese, Caiming Xiong, Ran Xu, Yejin Choi, and Ludwig Schmidt.
\newblock Mint-1t: Scaling open-source multimodal data by 10x: A multimodal dataset with one trillion tokens.
\newblock \emph{ArXiv}, abs/2406.11271, 2024.
\newblock URL \url{https://api.semanticscholar.org/CorpusID:270560059}.

\bibitem[Byeon et~al.(2022)Byeon, Park, Kim, Lee, Baek, and Kim]{kakaobrain2022coyo-700m}
Minwoo Byeon, Beomhee Park, Haecheon Kim, Sungjun Lee, Woonhyuk Baek, and Saehoon Kim.
\newblock Coyo-700m: Image-text pair dataset.
\newblock \url{https://github.com/kakaobrain/coyo-dataset}, 2022.

\bibitem[Chen et~al.(2024)Chen, Li, Dong, Zhang, Zang, Chen, Duan, Wang, Qiao, Lin, et~al.]{chen2024we}
Lin Chen, Jinsong Li, Xiaoyi Dong, Pan Zhang, Yuhang Zang, Zehui Chen, Haodong Duan, Jiaqi Wang, Yu~Qiao, Dahua Lin, et~al.
\newblock Are we on the right way for evaluating large vision-language models?
\newblock \emph{arXiv preprint arXiv:2403.20330}, 2024.

\bibitem[Fan et~al.(2023)Fan, Krishnan, Isola, Katabi, and Tian]{fan2023improving}
Lijie Fan, Dilip Krishnan, Phillip Isola, Dina Katabi, and Yonglong Tian.
\newblock Improving clip training with language rewrites.
\newblock In \emph{NeurIPS}, 2023.

\bibitem[Fang et~al.(2023)Fang, Jose, Jain, Schmidt, Toshev, and Shankar]{fang2023data}
Alex Fang, Albin~Madappally Jose, Amit Jain, Ludwig Schmidt, Alexander Toshev, and Vaishaal Shankar.
\newblock Data filtering networks.
\newblock \emph{arXiv preprint arXiv:2309.17425}, 2023.

\bibitem[Gadre et~al.(2023)Gadre, Ilharco, Fang, Hayase, Smyrnis, Nguyen, Marten, Wortsman, Ghosh, Zhang, Orgad, Entezari, Daras, Pratt, Ramanujan, Bitton, Marathe, Mussmann, Vencu, Cherti, Krishna, Koh, Saukh, Ratner, Song, Hajishirzi, Farhadi, Beaumont, Oh, Dimakis, Jitsev, Carmon, Shankar, and Schmidt]{datacomp}
Samir~Yitzhak Gadre, Gabriel Ilharco, Alex Fang, Jonathan Hayase, Georgios Smyrnis, Thao Nguyen, Ryan Marten, Mitchell Wortsman, Dhruba Ghosh, Jieyu Zhang, Eyal Orgad, Rahim Entezari, Giannis Daras, Sarah Pratt, Vivek Ramanujan, Yonatan Bitton, Kalyani Marathe, Stephen Mussmann, Richard Vencu, Mehdi Cherti, Ranjay Krishna, Pang~Wei Koh, Olga Saukh, Alexander Ratner, Shuran Song, Hannaneh Hajishirzi, Ali Farhadi, Romain Beaumont, Sewoong Oh, Alex Dimakis, Jenia Jitsev, Yair Carmon, Vaishaal Shankar, and Ludwig Schmidt.
\newblock Datacomp: In search of the next generation of multimodal datasets.
\newblock \emph{arXiv preprint arXiv:2304.14108}, 2023.

\bibitem[Kembhavi et~al.(2016)Kembhavi, Salvato, Kolve, Seo, Hajishirzi, and Farhadi]{Kembhavi2016ADI}
Aniruddha Kembhavi, Michael Salvato, Eric Kolve, Minjoon Seo, Hannaneh Hajishirzi, and Ali Farhadi.
\newblock A diagram is worth a dozen images.
\newblock \emph{ArXiv}, abs/1603.07396, 2016.
\newblock URL \url{https://api.semanticscholar.org/CorpusID:2682274}.

\bibitem[Lai et~al.(2023)Lai, Zhang, Wu, Bai, Timofeev, Du, Gan, Shan, Chuah, Yang, and Cao]{Lai2023VeCLIPIC}
Zhengfeng Lai, Haotian Zhang, Wentao Wu, Haoping Bai, Aleksei Timofeev, Xianzhi Du, Zhe Gan, Jiulong Shan, Chen-Nee Chuah, Yinfei Yang, and Meng Cao.
\newblock Veclip: Improving clip training via visual-enriched captions.
\newblock \emph{ArXiv}, 2023.
\newblock URL \url{https://api.semanticscholar.org/CorpusID:263835242}.

\bibitem[Laurenccon et~al.(2023)Laurenccon, Saulnier, Tronchon, Bekman, Singh, Lozhkov, Wang, Karamcheti, Rush, Kiela, Cord, and Sanh]{Laurenccon2023OBELISCAO}
Hugo Laurenccon, Lucile Saulnier, L{\'e}o Tronchon, Stas Bekman, Amanpreet Singh, Anton Lozhkov, Thomas Wang, Siddharth Karamcheti, Alexander~M. Rush, Douwe Kiela, Matthieu Cord, and Victor Sanh.
\newblock Obelisc: An open web-scale filtered dataset of interleaved image-text documents.
\newblock \emph{ArXiv}, abs/2306.16527, 2023.
\newblock URL \url{https://api.semanticscholar.org/CorpusID:259287020}.

\bibitem[Laurençon et~al.(2024)Laurençon, Tronchon, Cord, and Sanh]{Laurenon2024WhatMW}
Hugo Laurençon, L{\'e}o Tronchon, Matthieu Cord, and Victor Sanh.
\newblock What matters when building vision-language models?
\newblock \emph{ArXiv}, abs/2405.02246, 2024.
\newblock URL \url{https://api.semanticscholar.org/CorpusID:269587869}.

\bibitem[Li et~al.(2024{\natexlab{a}})Li, Zhang, Zhang, Guo, Zhang, Zhang, Li, Liu, and Li]{li2024llavanext-ablations}
Bo~Li, Hao Zhang, Kaichen Zhang, Dong Guo, Yuanhan Zhang, Renrui Zhang, Feng Li, Ziwei Liu, and Chunyuan Li.
\newblock Llava-next: What else influences visual instruction tuning beyond data?, May 2024{\natexlab{a}}.
\newblock URL \url{https://llava-vl.github.io/blog/2024-05-25-llava-next-ablations/}.

\bibitem[Li et~al.(2022)Li, Li, Xiong, and Hoi]{Li2022BLIPBL}
Junnan Li, Dongxu Li, Caiming Xiong, and Steven C.~H. Hoi.
\newblock Blip: Bootstrapping language-image pre-training for unified vision-language understanding and generation.
\newblock In \emph{International Conference on Machine Learning}, 2022.
\newblock URL \url{https://api.semanticscholar.org/CorpusID:246411402}.

\bibitem[Li et~al.(2024{\natexlab{b}})Li, Chen, Wang, Wang, Ye, Jin, Chen, He, Gao, Cui, Yu, Tian, Zhou, Xu, Wang, Wei, Li, Zhang, Zhang, Cai, Wen, Yan, Chu, Wang, Dou, Tian, Zhu, Lu, Chen, He, Lu, Wang, Wang, Lin, Qiao, Shi, He, and Dai]{Li2024OmniCorpusAU}
Qingyun Li, Zhe Chen, Weiyun Wang, Wenhai Wang, Shenglong Ye, Zhenjiang Jin, Guanzhou Chen, Yinan He, Zhangwei Gao, Erfei Cui, Jiashuo Yu, Hao Tian, Jiasheng Zhou, Chaochao Xu, Bin Wang, Xingjian Wei, Wei Li, Wenjian Zhang, Bo~Zhang, Pinlong Cai, Licheng Wen, Xiangchao Yan, Pei Chu, Yi~Wang, Min Dou, Changyao Tian, Xizhou Zhu, Lewei Lu, Yushi Chen, Jun-Jian He, Tong Lu, Yali Wang, Limin Wang, Dahua Lin, Yu~Qiao, Bo~Shi, Conghui He, and Jifeng Dai.
\newblock Omnicorpus: A unified multimodal corpus of 10 billion-level images interleaved with text.
\newblock \emph{ArXiv}, abs/2406.08418, 2024{\natexlab{b}}.
\newblock URL \url{https://api.semanticscholar.org/CorpusID:270391186}.

\bibitem[Li et~al.(2024{\natexlab{c}})Li, Tu, Hui, Wang, Zhao, Xiao, Ren, Mei, Liu, Zheng, Zhou, and Xie]{Li2024WhatIW}
Xianhang Li, Haoqin Tu, Mude Hui, Zeyu Wang, Bingchen Zhao, Junfei Xiao, Sucheng Ren, Jieru Mei, Qing Liu, Huangjie Zheng, Yuyin Zhou, and Cihang Xie.
\newblock What if we recaption billions of web images with llama-3?
\newblock \emph{ArXiv}, abs/2406.08478, 2024{\natexlab{c}}.
\newblock URL \url{https://api.semanticscholar.org/CorpusID:270391661}.

\bibitem[Liu et~al.(2023{\natexlab{a}})Liu, Li, Wu, and Lee]{Liu2023VisualIT}
Haotian Liu, Chunyuan Li, Qingyang Wu, and Yong~Jae Lee.
\newblock Visual instruction tuning.
\newblock \emph{ArXiv}, abs/2304.08485, 2023{\natexlab{a}}.
\newblock URL \url{https://api.semanticscholar.org/CorpusID:258179774}.

\bibitem[Liu et~al.(2023{\natexlab{b}})Liu, Duan, Zhang, Li, Zhang, Zhao, Yuan, Wang, He, Liu, Chen, and Lin]{Liu2023MMBenchIY}
Yuanzhan Liu, Haodong Duan, Yuanhan Zhang, Bo~Li, Songyang Zhang, Wangbo Zhao, Yike Yuan, Jiaqi Wang, Conghui He, Ziwei Liu, Kai Chen, and Dahua Lin.
\newblock Mmbench: Is your multi-modal model an all-around player?
\newblock \emph{ArXiv}, abs/2307.06281, 2023{\natexlab{b}}.
\newblock URL \url{https://api.semanticscholar.org/CorpusID:259837088}.

\bibitem[Liu et~al.(2024)Liu, Li, Huang, Yang, Yu, Li, Yin, lin Liu, Jin, and Bai]{liu2024ocrbenchhiddenmysteryocr}
Yuliang Liu, Zhang Li, Mingxin Huang, Biao Yang, Wenwen Yu, Chunyuan Li, Xucheng Yin, Cheng lin Liu, Lianwen Jin, and Xiang Bai.
\newblock Ocrbench: On the hidden mystery of ocr in large multimodal models, 2024.
\newblock URL \url{https://arxiv.org/abs/2305.07895}.

\bibitem[Lu et~al.(2022)Lu, Mishra, Xia, Qiu, Chang, Zhu, Tafjord, Clark, and Kalyan]{lu2022learn}
Pan Lu, Swaroop Mishra, Tony Xia, Liang Qiu, Kai-Wei Chang, Song-Chun Zhu, Oyvind Tafjord, Peter Clark, and Ashwin Kalyan.
\newblock Learn to explain: Multimodal reasoning via thought chains for science question answering.
\newblock In \emph{The 36th Conference on Neural Information Processing Systems (NeurIPS)}, 2022.

\bibitem[Masry et~al.(2022)Masry, Long, Tan, Joty, and Hoque]{Masry2022ChartQAAB}
Ahmed Masry, Do~Xuan Long, Jia~Qing Tan, Shafiq~R. Joty, and Enamul Hoque.
\newblock Chartqa: A benchmark for question answering about charts with visual and logical reasoning.
\newblock \emph{ArXiv}, abs/2203.10244, 2022.
\newblock URL \url{https://api.semanticscholar.org/CorpusID:247593713}.

\bibitem[Mathew et~al.(2021)Mathew, Bagal, Tito, Karatzas, Valveny, and Jawahar]{Mathew2021InfographicVQA}
Minesh Mathew, Viraj Bagal, Rub{\`e}n~P{\'e}rez Tito, Dimosthenis Karatzas, Ernest Valveny, and C.V. Jawahar.
\newblock Infographicvqa.
\newblock \emph{2022 IEEE/CVF Winter Conference on Applications of Computer Vision (WACV)}, pages 2582--2591, 2021.
\newblock URL \url{https://api.semanticscholar.org/CorpusID:233394125}.

\bibitem[Nguyen et~al.(2023)Nguyen, Gadre, Ilharco, Oh, and Schmidt]{Nguyen2023ImprovingMD}
Thao Nguyen, Samir~Yitzhak Gadre, Gabriel Ilharco, Sewoong Oh, and Ludwig Schmidt.
\newblock Improving multimodal datasets with image captioning.
\newblock \emph{ArXiv}, abs/2307.10350, 2023.
\newblock URL \url{https://api.semanticscholar.org/CorpusID:259991316}.

\bibitem[Schuhmann et~al.(2022)Schuhmann, Beaumont, Vencu, Gordon, Wightman, Cherti, Coombes, Katta, Mullis, Wortsman, Schramowski, Kundurthy, Crowson, Schmidt, Kaczmarczyk, and Jitsev]{Schuhmann2022LAION5BAO}
Christoph Schuhmann, Romain Beaumont, Richard Vencu, Cade Gordon, Ross Wightman, Mehdi Cherti, Theo Coombes, Aarush Katta, Clayton Mullis, Mitchell Wortsman, Patrick Schramowski, Srivatsa Kundurthy, Katherine Crowson, Ludwig Schmidt, Robert Kaczmarczyk, and Jenia Jitsev.
\newblock Laion-5b: An open large-scale dataset for training next generation image-text models.
\newblock \emph{ArXiv}, abs/2210.08402, 2022.
\newblock URL \url{https://api.semanticscholar.org/CorpusID:252917726}.

\bibitem[Singh et~al.(2019)Singh, Natarjan, Shah, Jiang, Chen, Batra, Parikh, and Rohrbach]{singh2019towards}
Amanpreet Singh, Vivek Natarjan, Meet Shah, Yu~Jiang, Xinlei Chen, Dhruv Batra, Devi Parikh, and Marcus Rohrbach.
\newblock Towards vqa models that can read.
\newblock In \emph{Proceedings of the IEEE Conference on Computer Vision and Pattern Recognition}, pages 8317--8326, 2019.

\bibitem[Wang et~al.(2023)Wang, Lv, Yu, Hong, Qi, Wang, Ji, Yang, Zhao, Song, Xu, Xu, Li, Dong, Ding, and Tang]{Wang2023CogVLMVE}
Weihan Wang, Qingsong Lv, Wenmeng Yu, Wenyi Hong, Ji~Qi, Yan Wang, Junhui Ji, Zhuoyi Yang, Lei Zhao, Xixuan Song, Jiazheng Xu, Bin Xu, Juanzi Li, Yuxiao Dong, Ming Ding, and Jie Tang.
\newblock Cogvlm: Visual expert for pretrained language models.
\newblock \emph{ArXiv}, abs/2311.03079, 2023.
\newblock URL \url{https://api.semanticscholar.org/CorpusID:265034288}.

\bibitem[Xue et~al.(2024)Xue, Shu, Awadalla, Wang, Yan, Purushwalkam, Zhou, Prabhu, Dai, Ryoo, Kendre, Zhang, Qin, Zhang, Chen, Yu, Tan, Awalgaonkar, Heinecke, Wang, Choi, Schmidt, Chen, Savarese, Niebles, Xiong, and Xu]{xgen_mm_phi3_mini}
Le~Xue, Manli Shu, Anas Awadalla, Jun Wang, An~Yan, Senthil Purushwalkam, Honglu Zhou, Viraj Prabhu, Yutong Dai, Michael~S Ryoo, Shrikant~B. Kendre, Jieyu Zhang, Can Qin, Shu~Zhen Zhang, Chia-Chih Chen, Ning Yu, Juntao Tan, Tulika Awalgaonkar, Shelby Heinecke, Huan Wang, Yejin Choi, Ludwig Schmidt, Zeyuan Chen, Silvio Savarese, Juan~Carlos Niebles, Caiming Xiong, and Ran Xu.
\newblock xgen-mm (blip-3): A family of open large multimodal models.
\newblock \emph{ArXiv}, 2024.
\newblock URL \url{https://api.semanticscholar.org/CorpusID:271891872}.

\bibitem[Yang et~al.(2024)Yang, Yang, Hui, Zheng, Yu, Zhou, Li, Li, Liu, Huang, Dong, Wei, Lin, Tang, Wang, Yang, Tu, Zhang, Ma, Xu, Zhou, Bai, He, Lin, Dang, Lu, Chen, Yang, Li, Xue, Ni, Zhang, Wang, Peng, Men, Gao, Lin, Wang, Bai, Tan, Zhu, Li, Liu, Ge, Deng, Zhou, Ren, Zhang, Wei, Ren, Fan, Yao, Zhang, Wan, Chu, Cui, Zhang, and Fan]{Yang2024Qwen2TR}
An~Yang, Baosong Yang, Binyuan Hui, Bo~Zheng, Bowen Yu, Chang Zhou, Chengpeng Li, Chengyuan Li, Dayiheng Liu, Fei Huang, Guanting Dong, Haoran Wei, Huan Lin, Jialong Tang, Jialin Wang, Jian Yang, Jianhong Tu, Jianwei Zhang, Jianxin Ma, Jin Xu, Jingren Zhou, Jinze Bai, Jinzheng He, Junyang Lin, Kai Dang, Keming Lu, Ke-Yang Chen, Kexin Yang, Mei Li, Min Xue, Na~Ni, Pei Zhang, Peng Wang, Ru~Peng, Rui Men, Ruize Gao, Runji Lin, Shijie Wang, Shuai Bai, Sinan Tan, Tianhang Zhu, Tianhao Li, Tianyu Liu, Wenbin Ge, Xiaodong Deng, Xiaohuan Zhou, Xingzhang Ren, Xinyu Zhang, Xipin Wei, Xuancheng Ren, Yang Fan, Yang Yao, Yichang Zhang, Yunyang Wan, Yunfei Chu, Zeyu Cui, Zhenru Zhang, and Zhi-Wei Fan.
\newblock Qwen2 technical report.
\newblock \emph{ArXiv}, abs/2407.10671, 2024.
\newblock URL \url{https://api.semanticscholar.org/CorpusID:271212307}.

\bibitem[Yu et~al.(2023)Yu, Sun, Zhang, Cui, Zhang, Wang, and Liu]{Yu2023CapsFusionRI}
Qiying Yu, Quan Sun, Xiaosong Zhang, Yufeng Cui, Fan Zhang, Xinlong Wang, and Jingjing Liu.
\newblock Capsfusion: Rethinking image-text data at scale.
\newblock \emph{ArXiv}, abs/2310.20550, 2023.
\newblock URL \url{https://api.semanticscholar.org/CorpusID:264828939}.

\bibitem[Zhai et~al.(2023)Zhai, Mustafa, Kolesnikov, and Beyer]{Zhai2023SigmoidLF}
Xiaohua Zhai, Basil Mustafa, Alexander Kolesnikov, and Lucas Beyer.
\newblock Sigmoid loss for language image pre-training.
\newblock \emph{2023 IEEE/CVF International Conference on Computer Vision (ICCV)}, pages 11941--11952, 2023.
\newblock URL \url{https://api.semanticscholar.org/CorpusID:257767223}.

\bibitem[Zhang et~al.(2024)Zhang, Li, Zhang, Pu, Cahyono, Hu, Liu, Zhang, Yang, Li, and Liu]{zhang2024lmmsevalrealitycheckevaluation}
Kaichen Zhang, Bo~Li, Peiyuan Zhang, Fanyi Pu, Joshua~Adrian Cahyono, Kairui Hu, Shuai Liu, Yuanhan Zhang, Jingkang Yang, Chunyuan Li, and Ziwei Liu.
\newblock Lmms-eval: Reality check on the evaluation of large multimodal models, 2024.
\newblock URL \url{https://arxiv.org/abs/2407.12772}.

\bibitem[Zhang et~al.(2023)Zhang, Zhang, Gu, Zhou, Lipka, Yang, and Sun]{Zhang2023LLaVAREV}
Yanzhe Zhang, Ruiyi Zhang, Jiuxiang Gu, Yufan Zhou, Nedim Lipka, Diyi Yang, and Tongfei Sun.
\newblock Llavar: Enhanced visual instruction tuning for text-rich image understanding.
\newblock \emph{ArXiv}, abs/2306.17107, 2023.
\newblock URL \url{https://api.semanticscholar.org/CorpusID:259287523}.

\bibitem[Zhu et~al.(2023)Zhu, Hessel, Awadalla, Gadre, Dodge, Fang, Yu, Schmidt, Wang, and Choi]{Zhu2023MultimodalCA}
Wanrong Zhu, Jack Hessel, Anas Awadalla, Samir~Yitzhak Gadre, Jesse Dodge, Alex Fang, Youngjae Yu, Ludwig Schmidt, William~Yang Wang, and Yejin Choi.
\newblock Multimodal c4: An open, billion-scale corpus of images interleaved with text.
\newblock \emph{ArXiv}, abs/2304.06939, 2023.
\newblock URL \url{https://api.semanticscholar.org/CorpusID:258170467}.

\end{thebibliography}
